# Diagnosis of Human–Object Interaction Detectors for Real-World Educational Applications

Divya Mereddy Ashwin Tudur Sadashiva Marcos Quiñones-Grueiro Gautam Biswas
Vanderbilt University
Nashville, TN, USA
{divya.mereddy , ashwin.tudur.sadashiva, marcos.quinones.grueiro, gautam.biswas }
@vanderbilt.edu

## Abstract

*Human–object interaction (HOI) recognition is critical for automatically analyzing student behavior in complex educational environments. Although state-of-the-art (SOTA) HOI detectors perform well on benchmark datasets, their performance often degrades when deployed in real-world training environments due to domain-specific objects, occlusions, and complex visual conditions. In this paper, we introduce a diagnosis-driven framework that integrates a triplet-level HOI error taxonomy with error–factor attribution analysis for real-world educational video data. We study this problem in the context of Critical Care Air Transport Team (CCATT) mixed-reality medical training. Based on an analysis of HOI failure modes and their causes, we develop a diagnosis-informed refinement strategy for adapting pretrained HOI models to the target domain. Experiments on the CCATT dataset show that this approach improves the macro-F1 score of a pretrained CDN model from 48.6 to 90.2 through targeted refinement guided by diagnosed error factors. These results highlight the value of detailed diagnostic analysis for informing targeted adaptation of HOI models in real-world educational environments.*

## 1. Introduction

Action recognition and human–object interaction (HOI) analysis provide an important mechanism for understanding how learners engage with task-relevant artifacts in complex learning and training environments [9, 14, 22]. In educational and simulation-based training scenarios, many meaningful behaviors are expressed through interactions with equipment and tools, making HOI recognition a critical component for automated behavioral analysis and student assessment system. HOI-based observable interactions serve as measurable proxies for latent cognitive processes and support automated trainee evaluation systems in medical training simulations such as the Critical Care Air Transport Team (CCATT) training environment [19]. The CCATT environment is an en-route medical care setting in which a three-member critical care medical team is responsible for stabilizing and managing critically injured patients during transport. To support trainee evaluation in CCATT teams with dynamically changing trainee composition and patient counts, where manual analysis is limited, such automated AI-based assisting systems are essential.

Recent state-of-the-art human–object interaction (HOI) detectors (e.g., CDN [25], PVIC [26], and HOICLIP [20]) support object-centric action recognition through human–object interaction triplets and achieve strong performance on standard benchmark datasets. However, models trained under these benchmark conditions often fail to generalize to real-world, in-the-wild educational environments such as CCATT, where visual conditions and interaction patterns differ substantially from those observed in curated datasets. In particular, heterogeneous lighting and imaging conditions—such as grayscale, RGB, headlamp-only lighting, and low-light settings—together with motion blur and occlusion, limited domain-specific knowledge, and the increased complexity of HOI scenes (e.g., multiple interactions occurring simultaneously and visually similar objects and persons in close proximity) make reliable deployment of pretrained HOI architectures challenging in real-world educational environments. Moreover, benchmark HOI datasets are often biased toward positive interaction instances, which can lead to systematic false positives in educational settings where near-contact and incidental proximity are common but should be interpreted as no-interaction. Consequently, substantial refinement is often required to enable models to reliably differentiate valid interactions from non-interaction scenarios.

Diagnostic toolboxes and error-analysis frameworks have proven effective in accelerating progress on standard computer vision tasks, including HOI, by providing fine-grained quantitative analyses of model failures [6, 11]. However, best practices for diagnosing HOI models in real-world educational data remain underexplored. By integrating a triplet-level HOI error taxonomy with computer vision error–factor attribution, this paper introduces a diagnostic framework to quantify and interpret the failure modes of modern HOI detectors and the conditions that co-occur with them in real-world educational settings. Based on this analysis, we identify targeted adaptation strategies for improving HOI model performance on educational data characterized by complex visual conditions and interaction variability.

Overall, the main contributions of this paper are as follows:

- We introduce a diagnostic framework for analyzing the performance of human–object interaction (HOI) models in real-world educational environments. We leverage a triplet-level HOI error taxonomy and apply it to real-world educational video data, integrating it with error–factor attribution to analyze model failure modes under deployment conditions.
- We propose diagnosis-guided refinement strategies for adapting pretrained HOI models to target datasets, informed by model error analysis. We further demonstrate this process in the CCATT setting.
- We conduct an error analysis of pretrained HOI detectors in a mixed-reality en-route critical care training environment (CCATT) and show how the resulting diagnostic insights informed the adaptation of a pretrained CDN model.

# 2. Background

## 2.1. Learning Environment and Data Description

CCATT is a specialized U.S. Air Force medical unit responsible for stabilizing and transporting critically injured soldiers during aeromedical evacuation [13]. To prepare personnel for in-flight critical care, CCATT mixed-reality training is designed to approximate the operational, environmental, and cognitive demands of aircraft-based care (Figure 1). A standard CCATT team comprises a physician, a nurse, and a respiratory therapist.

The initial CCATT simulation features two CCATT teams, each with four manikin-based patient beds, each equipped with critical care devices such as mechanical ventilators, IV infusion equipment, and ProPaq monitors. Teams of three trainees manage one or two patients simultaneously. Vision data quality suffers due to frequent trainee occlusions of devices and peers. Lighting varies to simulate in-flight conditions, including low-light headlamp use, infrared grayscale for instructors, and RGB footage. Our dataset comprises 20 initial training sessions, each recorded with multiple cameras. But in this study we utilized three cameras data($\sim$45 hours data with $\sim$3.5 million frames sampled at 4 FPS) for error analysis and two cameras data($\sim$30 hours data with $\sim$3 million frames sampled at 4 FPS) for fine-tuning. We focus on interactions with IV equipment, Mechanical Ventilators (MV), and ProPaq monitors, labeling equipment engagement as “valid-interaction” and non-engagement as “no-interaction”, making a total of six types of Trainee-equipment interactions. Equipment engagement is considered a valid interaction only when the trainee actively manipulates device controls (e.g., adjusting settings, pressing buttons, silencing alarms) or performs actions necessary to acquire clinically relevant information from the device. Incidental proximity or contact without operational intent—is explicitly labeled as no-interaction.

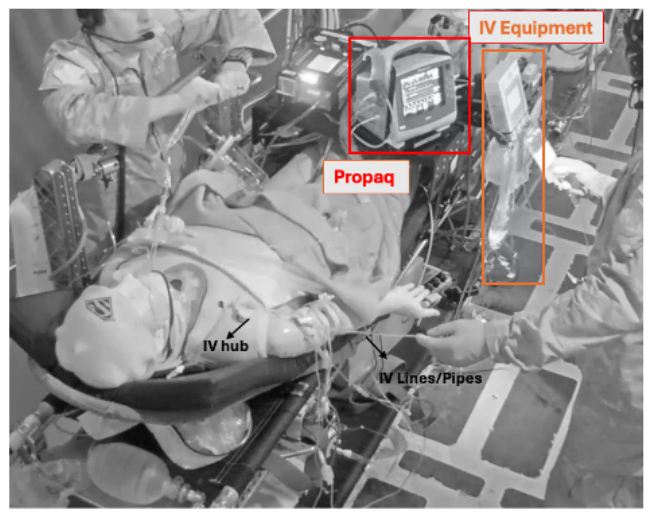

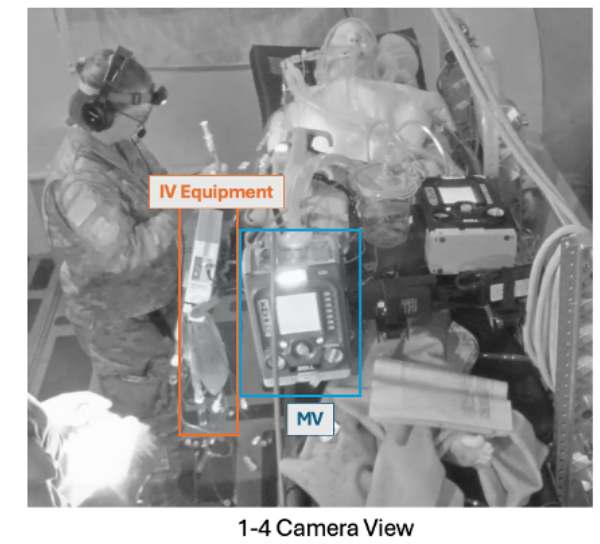

Figure 1. CCATT Video Data Multi Camera Views (Source: AFRL, Dayton)

## 2.2. Literature on HOI Detector Diagnosis

Model diagnosis and error analysis have become increasingly important in computer vision, as aggregate metrics alone often fail to explain why modern deep models break under deployment conditions [4, 6, 11, 28, 29]. This challenge is particularly pronounced in human–object interaction (HOI) detection, where a single prediction depends jointly on human localization, object localization, pair association, and predicate recognition. Recent work on diagnosing HOI detectors has shown that standard evaluation metrics such as mAP are insufficient for understanding model behavior and has introduced more fine-grained diagnostic pipelines for analyzing HOI error types [29].

On the other hand, in computer vision, several studies analyze factors that contribute to model errors, such as the effects of domain shift, variations in visual conditions (e.g., occlusion and illumination), and the adaptation of models to new application domains with previously unseen objects and actions [2, 5, 10, 15, 27]. Within HOI research, some papers analyze conditions that influence HOI errors [7, 23, 26]; however, systematic error–factor attribution frameworks remain relatively underexplored in the HOI literature. Existing HOI diagnostic and error analysis studies have been developed primarily in the context of benchmark datasets such as HICO-DET and V-COCO.

In instructional and training environments, the use of HOI-based student action recognition is growing [16, 19]. In these applications, HOI detectors must operate under substantially more complex conditions, including partial occlusion, multi-object and multi-person interactions, visually similar objects in close proximity, heterogeneous sensing and imaging settings (e.g., IR/low-light/headlamp scenarios), uncommon camera angles, and motion blur [4, 19].While prior work has proposed diagnostic taxonomies for HOI detection and separately analyzed dataset characteristics affecting model performance, existing studies typically analyze these aspects independently and do not explicitly integrate structured HOI error taxonomies with systematic error–factor attribution. Our framework bridges this gap by systematically linking triplet-level HOI error categories with associated contributing factors specially in the context of real-world educational data.

# 3. Methods

When deployed in real-world educational environments, pretrained HOI models often experience substantial performance degradation due to domain shift and the complexity of real-world data [3, 4, 19]. These challenges arise from multi-person and multi-object interactions, visually similar objects in close spatial proximity, heterogeneous visual conditions across imaging settings, and the presence of domain-specific entities such as specialized medical equipment (e.g., IV pumps and ventilators) that are absent from benchmark datasets. Consequently, zero-shot deployment often produces noisy predictions and systematic mislabeling, as models remain biased toward object categories observed during pretraining.

## 3.1. Error Analysis Framework

To understand why state-of-the-art HOI detectors fail in real-world educational video conditions and to guide improvements in prediction accuracy, we develop a diagnostic framework for analyzing the performance of human–object interaction (HOI) models in real-world educational environments. The proposed framework adopts a triplet-level HOI error taxonomy and integrates it with an error–factor attribution analysis tailored to real-world educational video data. This integration enables systematic interpretation of failure modes under real-world deployment conditions.

Based on the diagnostic insights obtained from this analysis, we further develop diagnosis-guided refinement strategies to adapt pretrained HOI models to custom datasets. These strategies leverage the identified error categories and associated factors to guide targeted model adaptation in the CCATT environment. We quantify the observed errors (and co-occurring error factors) through manual error bucketing on prediction cases where the ground-truth labels are unambiguous to human annotators as shown in Figure 2.

### 3.1.1. Triplet-level taxonomy.

Following [29], given a model prediction in triplet form $\langle h, v, o\rangle$, we assign each predicted triplet to exactly one mutually exclusive error bucket using the logical flow in Figure 3). This decision procedure aligns with the two coupled sub-tasks of HOI detection—(i) human–object pair localization/association and (ii) predicate recognition—and yields a consistent accounting of failure modes across models and recording conditions. Concretely, we categorize errors into *Detection Error–Object*, *Detection Error–Person*, *Association Error* (including multi-person confusion), *Both Boxes Error*, *Action Error* (verb/triplet misclassification), and *Duplicate Error*. Each bucket is defined by the earliest point of failure in the HOI pipeline (e.g., localization vs. association vs. predicate), enabling error summaries that are directly actionable for model refinement.

### 3.1.2. Error–Factor Attribution

After error bucketing, we analyze which factors co-occur with each error type using Table 1. The factor columns are non-exclusive because a single failure may be influenced by multiple conditions. Errors are quantified through manual inspection of model predictions and assignment to the taxonomy defined in Sec. 3.1.1. We group contributing factors into two categories: *Visual degradation factors* and *domain mismatch*.

**Visual degradation factors:** Blur degrades visual features and can affect multiple stages of the HOI pipeline, leading to object detection, human detection, association, and action prediction errors [10]. Variations in illumination and sensing modality introduce new challenges. Illumination degradation (e.g., low-light conditions) reduces contrast and signal quality, whereas modality changes (e.g., infrared imaging) introduce a distribution shift relative to RGB-trained models [12, 18]. Both factors can lead to missed detections or incorrect interaction predictions. Occlusions frequently arise in complex environments where multiple individuals operate in confined spaces, obscuring critical visual cues required for detection and interaction recognition, and resulting in bounding-box localization errors or incorrect human–object associations. Camera-angle changes alter the visual appearance of objects and actions; when interactions are observed from uncommon viewpoints, pretrained models trained on benchmark datasets may fail to recognize them correctly [1]. Crowded scenes further increase scene density and human–object pairing ambiguity. When multiple humans or visually similar objects appear in close proximity, HOI detectors may associate interactions with incorrect instances, producing association errors.

**Domain mismatch:** Domain mismatch occurs when the deployment environment differs from the pretraining distribution. Domain-specific objects absent from benchmark datasets can produce object detection errors and downstream HOI failures. Differences in human appearance, such as mannequins, specialized uniforms, or children in educational settings, may also affect person detection. Domain mismatch may also occur at the action level when target-domain interactions are not represented in the source dataset, resulting in incorrect verb predictions.

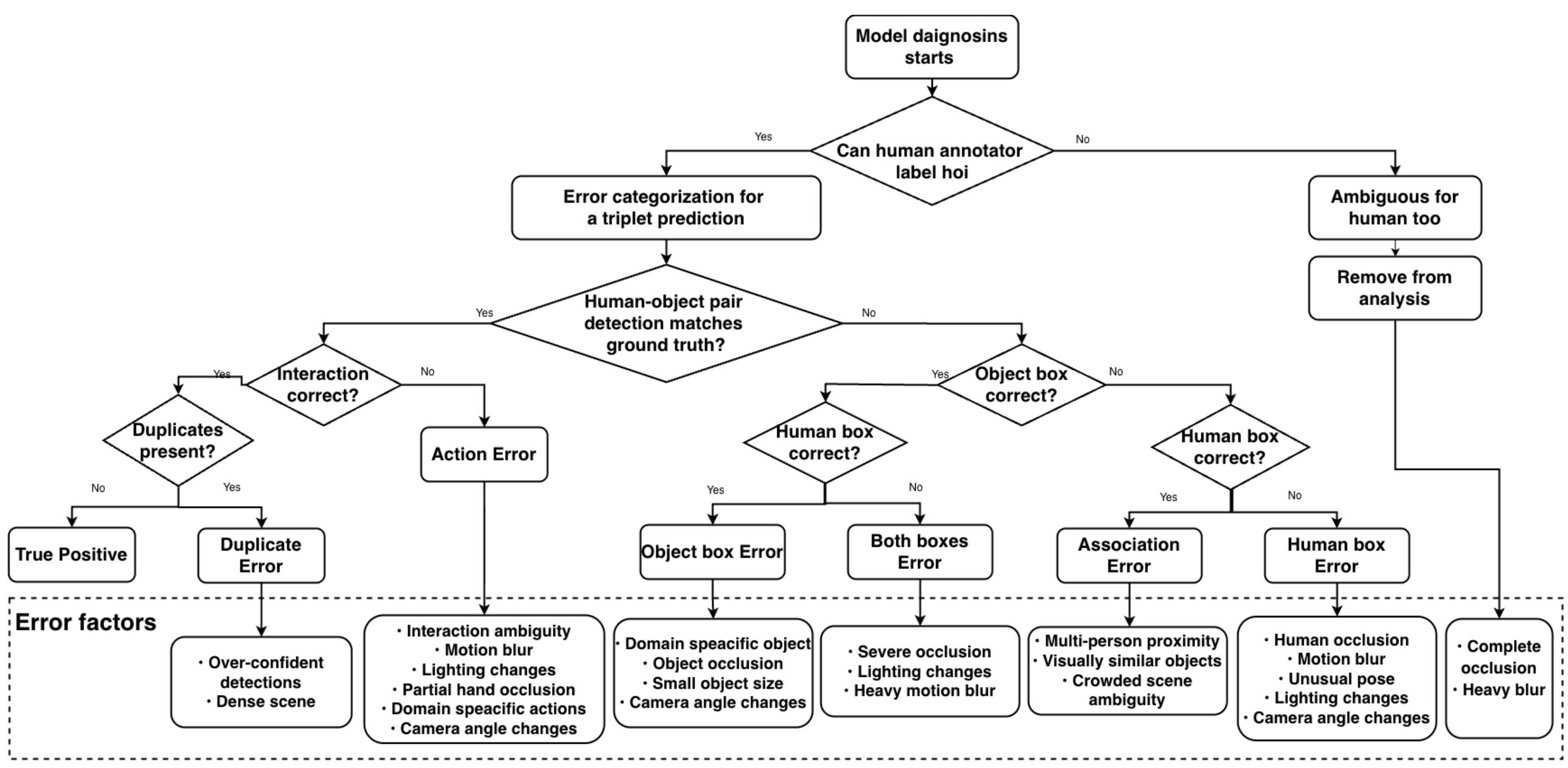

Figure 2. Error categorization flow. Note that, different types of errors are mutually exclusive

Table 1. Error–factor matrix with grouped factors(quantifiable). Columns are non-exclusive; each colored checkmark indicates the primary mitigation: ✓Automated refinements, ✓Human annotator-in-the-loop refinement, ✓Not addressable under current constraints.

| HOI Error Bucket | Visual Degradation Factors | | | | | Domain Mismatch | | |
|---|---|---|---|---|---|---|---|---|
| | Low-light /IR | Blur | Semi–Occlusion | Camera angle changes | Crowded Scenes | Objects | Actions | Humans |
| Detection – Object Box Error | ✓ | ✓ | ✓ | ✓ | ✓ | ✓ | | |
| Detection – Human Box Error | ✓ | ✓ | ✓ | ✓ | ✓ | | | ✓ |
| Detection – Both Boxes Error | ✓ | ✓ | ✓ | ✓ | ✓ | ✓ | | |
| Association Error | | ✓ | ✓ | ✓ | ✓ | | | |
| Action Error | ✓ | ✓ | ✓ | ✓ | | | ✓ | |
| Duplicate Error | | | | | | | | |

The factors summarized in Table 1 correspond to environmental and domain conditions influencing model perception during inference and are quantified through manual error bucketing on predictions from pretrained model inference. In contrast, Table 2 describes factors arising from dataset characteristics and pretraining biases, which are analyzed through dataset statistics and annotation-level inspection rather than architectural experiments.

**Data factors:**

**Class imbalance:** Long-tailed interaction distributions can bias models toward majority classes, reducing recall for minority interactions.

**Verb label ambiguity.** If there is any disagreement between human annotators as well, we ignored them as “ambiguous/uncertain” [7].

**Pretrained model limitations:** Biases learned during pretraining may persist during adaptation. Benchmark HOI datasets such as HICO-DET often under-represent explicit negative interaction categories (e.g., *no-interaction*) [7, 26]. As a result, pretrained models can exhibit a bias toward predicting interactions in near-contact or visually ambiguous scenarios.

**Verb label uncertainty/ambiguity:** If there is disagree-ment between annotators, the corresponding verb annotations are marked as "ambiguous/uncertain." Such cases are observed in a small fraction of HOI datasets [7] and typically arise due to inherent ambiguity in interpreting interactions from visual data. For example, an interaction annotated as "straddle bike" may also be labeled as "ride bike." Similarly, in the CCATT context, when an interaction has just begun and physical contact is minimal, such cases may be ambiguously labeled as either valid interaction or no-interaction.

Table 2. Error–factor matrix with grouped factors(distributive). Columns are non-exclusive; each colored check-mark indicates the primary mitigation: ✓Automated refinements, ✓Human annotator-in-the-loop refinement, ✓Not addressable under current constraints.

| HOI Error Bucket | Data Factors | | Pretrained Model Limitations |
|---|---|---|---|
| | Imbalance | Verb label ambiguity | No-interaction sparsity |
| Detection – Object Box Error | ✓ | | ✓ |
| Detection – Human Box Error | | | ✓ |
| Detection – Both Boxes Error | | | ✓ |
| Association Error | | | |
| Action Error | ✓ | ✓ | ✓ |
| Duplicate Error | | | |

Duplicate may arise from model overconfidence or sub-optimal inference thresholds selected under domain shift. Within the proposed framework, factor attribution identifies conditions that co-occur with specific failure modes based on the error bucketing analysis. The resulting error distributions motivate the targeted adaptation strategies described in Sec. 3.2.

**Ambiguous for human:** Some HOI instances exhibit inherent annotation ambiguity due to limited visual evidence, such as severe blur, complete occlusions or partial visibility.This ambiguity is further amplified in temporally transitional phases, where trainee–object interactions are in the process of initiation or termination. Instances that are ambiguous even for human annotators are excluded from the analysis.

## 3.2. Diagnosis of Errors for Robust Models

Based on the diagnostic analysis in Sec. 3.1, we develop targeted refinement strategies to improve the robustness of pretrained HOI models in the CCATT environment while limiting additional manual annotation. Rather than relying on large-scale annotated data based fine-tuning, the proposed approach adopts diagnosis-guided refinements, where mitigation strategies are selected according to the error causes identified through the error taxonomy and factor attribution analysis. To reduce manual effort while improving model robustness, the refinement process consists of two complementary strategies: (1) *automated refinement*, where improvements are achieved using AI-based methods such as data augmentation, automated bounding-box generation, or pseudo-labeling; and (2) *human-annotator-in-the-loop refinement*, where interaction labels are corrected through targeted human annotation following automated data labeling. Each mitigation strategy is selected based on the diagnosed error causes in Sec. 3.1.

**Visual degradation factors:**

**Blur:** To improve robustness under blur-related failures, we introduce blur-oriented augmentation during fine-tuning (*augmentation-based refinement*). Training samples are augmented using motion-blur and Gaussian-blur transformations.

**Lighting and imaging variations:** To address failures associated with heterogeneous imaging conditions, we introduce grayscale conversion and simulated illumination variations during training (*augmentation-based refinement*). These augmentations expose the model to diverse sensing conditions and improve robustness to distribution shifts between training and deployment environments.

**Occlusion:**To mitigate occlusion-related failures, we apply partial occlusion augmentation by masking portions of object and human regions during training (*augmentation-based refinement*). This improves robustness under incomplete visual observations.

**Crowded scenes:** For association errors in crowded scenes, we incorporate refinement data through a *human-annotator-in-the-loop* process. Ambiguous interaction predictions are reviewed and corrected by annotators, and the refined samples are incorporated into fine-tuning dataset.

**Camera-angle changes:** To address viewpoint-related failures, we incorporate a *human-annotator-in-the-loop* strategy. Mislabeled verb annotations are identified through manual inspection and subsequently corrected as part of the refinement process.

**Domain mismatch:**

**Objects and persons:** To address detection and classification errors caused by domain mismatch in object and human appearance, we employ domain-specific bounding-box detectors, including open-vocabulary detectors such as Grounding Dino[17], YOLO-World [8], to generate candidate bounding boxes for objects and persons. These automatically generated proposals are used to construct refinement training data with tighter and more reliable annotations. The generated boxes are incorporated into the fine-tuning dataset as part of an automated AI-assisted refinement strategy, enabling the HOI model to adapt to CCATT-specific object categories and human appearances while minimizing manual annotation.

**Actions:** To address interaction prediction errors associated with domain-specific actions, we introduce additional interaction labels relevant to the CCATT environment and refine them using human supervision (*human-annotator-in-the-loop*). Pretrained HOI models can leverage transferable verb representations learned from benchmark datasets to support learning of new interaction categories with limited supervision [24, 25].

**Data factor:**

**Class imbalance.** To address class imbalance and long-tailed interaction distributions identified during dataset analysis, we apply targeted augmentation to increase representation of minority interaction classes (*augmentation-based refinement*). In addition, imbalance-aware loss functions such as focal loss are used to emphasize hard and minority samples during optimization [21].

**Verb label uncertainty/ambiguity:** Within the proposed diagnostic framework, these instances are categorized as label ambiguity and are treated as non-actionable errors that cannot be reliably resolved. These cases are excluded from the training data to reduce ambiguity in model training process.

**Pretrained model limitations: No-interaction sparsity:** To mitigate biases inherited from benchmark HOI datasets that under-represent explicit negative interactions, we include *no-interaction* annotations during fine-tuning. These negative examples help the model distinguish valid interactions from spatial proximity between humans and objects and reduce false-positive interaction predictions during inference. *No-interaction* samples can be generated using pretrained model outputs. For each human–object pair without a valid interaction is labeled as a *no_interaction* HOI triplet, and verb annotations are manually verified when the overlap of human–object bounding boxes exceeds a predefined threshold.

Duplicate interaction predictions are typically mitigated through improved fine-tuning data and training procedures.

## 4. Diagnosis Results:

### 4.1. Setup:

We evaluate three representative state-of-the-art HOI detectors—CDN, HOICLIP, and PViC—to characterize failure modes and robustness under deployment-oriented conditions. We initialize them from publicly available pretrained checkpoints trained on the HICO-DET. We then analyzed the error distributions across models using the taxonomy and factor attribution framework introduced in Sec. 3. We manually analyzed randomly selected 2000 images for error analysis. For model fine-tuning we used 70% of data for training, 15% validation, 15% for testing, and concentrated on CDN one of the HOI models that we evaluated in error analysis. We perform dataset splitting at the session level, ensuring that all frames from a given session are assigned exclusively to a single split. In our analysis, we define infrared/grayscale and headlamp-only low-light scenes as illumination & modality-change mismatches, as HICO-DET are primarily contains RGB imagery.

Because the evaluated HOI detectors are pretrained on HICO-DET and are not exposed to CCATT-specific equipment categories, object category predictions can be unreliable under domain shift even when the predicted object region overlaps the target equipment [7]. To disentangle localization quality from category mismatch, we perform class-agnostic equipment assignment by matching predicted object boxes to predefined CCATT equipment regions using Intersection-over-Union (IoU). Following [19], if the maximum IoU exceeds a threshold $\tau$, the predicted box is assigned to the corresponding equipment region regardless of its predicted object label. We then analyze predicate scores conditioned on the matched equipment region and project a small set of semantically aligned benchmark verbs to CCATT interaction labels, treating these outputs as proxy interaction signals rather than supervised CCATT recognition. We considered both no-interaction predictions and HO pairs without any interaction prediction as CCATT no-interaction HOI prediction.

### 4.2. Error Analysis and Model Diagnosing Results

We quantify model failure modes using the mutually exclusive triplet-level taxonomy introduced in Sec. 3.1 and report the distribution of observed error buckets for three pretrained HOI detectors in Table 3. Because these models are initialized from benchmark pretraining on HICO-DET and are not exposed to CCATT-specific equipment categories, object predictions can be unreliable under domain shift even when the predicted object region overlaps the target equipment. The reported bucket distribution therefore reflects the failure modes observable under the evaluated deployment conditions rather than all potential factors defined in the diagnostic framework. Some factors may not appear if the source and target distributions are similar for those aspects. We therefore interpret the error distribution as a characterization of how pretrained HOI pipelines fail in the CCATT environment and use Table 3 to identify the most prominent error patterns for targeted model refinement.

For the CDN model, Table 3 shows that several errors co-occur with domain mismatch between the pretrained model and the CCATT environment. In particular, object-side domain mismatch (35.50%) frequently appears in detection errors, indicating difficulty in recognizing CCATT-specific equipment that is absent from the pretrained HOI datasets. Due to the limited prior knowledge of CCATT objects, object bounding boxes may be localized to some extent, but incorrect object class assignments are frequently observed. At the interaction level, action prediction errors co-occurring with domain mismatch (23.74%) and action classification errors (12.61%) remain substantial. The model often predicts generic proxy verbs such as hold, carry, or watch, reflecting predicate vocabulary mismatch between the pretrained HOI datasets (e.g., HICO-DET) and the CCATT task context. In some cases, the model defaults to the background predicate no_interaction, which is shared between the source benchmark dataset and the target environment.

Table 3. Distribution of HOI error buckets and associated co-occurring error factors across pretrained detectors (% of analyzed error cases).

| HOI Errors | Co-occurring Error Factors | CDN | PVIC | HOICLIP |
|---|---|---|---|---|
| Detection Object Box Error | Domain Mismatch | 35.50 | 38.45 | 2.10 |
| | Classification Error – Domain Mismatch | 100.00 | 100.00 | 100.00 |
| | Semi-occlusions | 3.99 | 3.36 | 0.84 |
| | Low-light / IR | 38.24 | 38.45 | 2.10 |
| Detection Person Box Error | Domain Mismatch | 1.05 | 1.05 | 5.25 |
| | Low-light / IR | 1.05 | 1.05 | 2.52 |
| | Semi-occlusions | 1.05 | 0.00 | 2.94 |
| Association Error | Crowded Scenes | 0.84 | 2.10 | 6.09 |
| | Low-light / IR | 0.63 | 2.10 | 6.30 |
| Action Error | Domain Mismatch | 23.74 | 21.01 | 8.61 |
| | Classification Error – Domain Mismatch | 12.61 | 17.86 | 22.48 |
| | Blurred Regions | 0.42 | 0.21 | 0.42 |
| | Low-light / IR | 13.03 | 17.86 | 8.82 |

A considerable portion of errors also co-occurs with visual challenges, including heterogeneous lighting conditions, motion blur, and partial occlusions, which degrade feature quality for both object detection and interaction recognition. Furthermore, crowded scenes introduce human–object pairing ambiguity, leading to association errors when multiple trainees interact with nearby equipment. Similar patterns can be observed for PVIC and HOICLIP. In particular, HOICLIP demonstrates lower object detection errors, as the its zero-shot boosting mechanism enables the model to localize object bounding boxes more accurately and with higher confidence. Representative qualitative examples of CCATT-specific errors grouped by error bucket are provided in the supplementary material and are available online at https://drive.google.com/drive/folders/176JCBn507VMx8hVcqmS4FTqAyfwgMx94?usp=sharing

Table 4. Performance comparison - pretrained Vs adapted CDN models.

| Model | Total Frames | Human Annotated | Marco-F1 |
|---|---|---|---|
| CDN (Pretrained) | – | – | 48.6 |
| CDN (Fine-tuned) | 7,650 | 1,828 (24%) | 90.2 |

### 4.3. Diagnosis-Guided CDN Adaptation for CCATT

Based on Sec. 4.2, we refine the pretrained CDN model through proposed diagnosis-guided adaptation procedure designed. the refinement process targets specific weaknesses, including object & verb domain mismatch, missed person detections,association ambiguity in multi-person scenes etc.

To address object-side domain mismatch, we first use pretrained model predictions to generate candidate interaction annotations as suggested in methods. CDN and skeleton based methods for verb predictions. Further, We incorporate human-in-the-loop annotation correction to revise verb labels and resolve association ambiguities, particularly in crowded scenes where multiple trainees appear in close proximity to the same equipment. This manual correction stage is important for preventing propagation of association errors into the refined training set.

The diagnostic analysis also showed that the pretrained CDN model does not explicitly model no-interaction, even though this distinction is critical in CCATT, where subtle differences separate operational engagement from incidental proximity. We therefore introduce explicit no-interaction annotations during refinement as discussed in [19]. Including these negative examples improves boundary formation between interaction and non-interaction cases and reduces action-level confusion in subsequent adaptation. During the initial analysis in Sec. 4.2, class imbalance effects were not clearly observable because many source HOI objects and predicates in the pretrained models do not match the target CCATT interaction distribution. However, during the refinement process we observed imbalance across target dataset distribution for interactions involving for Propaq monitors. We therefore incorporated imbalance handling when constructing the fine-tuning dataset.

Finally, to improve robustness to deployment-specific sensing conditions, we re-fine-tune the model using the corrected annotations together with targeted augmentations that simulate low-light conditions, illumination variation, blur, and partial occlusion. These augmentations are intended to address the visibility-related failure modes identified in Sec. 4.2 and to improve robustness to conditions that differ from those represented in pretrained HOI data. Overall, this diagnosis-guided refinement procedure incorporates automated and human-annotator-in-the-loop approaches to construct a targeted fine-tuning dataset for adapting pretrained models to the CCATT educational setting.

After applying the proposed refinement pipeline, the adapted CDN model improves over the pretrained baseline, with macro-F1 increasing from 48.6 to 90.2 [Table 4]. To limit direct expert annotation effort, we adopt the SAAL-HOI (Semi-Automatic Active Learning for HOI) pipeline. Rather than treating adaptation as a one-time large-scale annotation process, SAAL-HOI formulates refinement as an iterative procedure that combines selective human verification (active learning) with automatic pseudo-label expansion (self-training), thereby avoiding unnecessary manual labeling of easy samples. Using this approach, we construct a custom annotated dataset of 7,650 frames, of which only 24% (1,828 frames) require direct human labeling. Despite this limited supervision, the adapted CDN model showed a large improvement over the pretrained baseline, suggesting that targeted fine-tuning can support domain adaptation in the CCATT setting with limited direct human labeling.

# 5. Discussion and Conclusions

This work investigates the reliability of pretrained human–object interaction detectors in real-world educational training environments. We introduce a diagnosis-guided framework that combines a triplet-level HOI error taxonomy with error–factor attribution analysis and show how these insights informed targeted refinement strategies in the CCATT deployment setting.

Our empirical analysis in the CCATT mixed-reality training environment reveals consistent failure patterns in pretrained HOI detectors, with dominant errors arising from predicate misclassification, object-side domain mismatch, and ambiguity caused by occlusion and multi-person proximity. Guided by these findings, we apply diagnosis-guided refinements that enable efficient domain adaptation of pretrained models while minimizing additional manual annotation effort.

Although this study refines the CDN architecture, Vision–language HOI models such as HOICLIP can offer advantages when the target environment introduces substantial semantic variation or requires zero-shot interaction generalization. In contrast, when the target distribution remains relatively similar to the fine-tuning data distribution, as in the current CCATT setting, a vision-based architecture such as CDN can remain effective. As this study is focused on a single real-world educational dataset (CCATT), and therefore the generalization of the proposed framework to other application domains may require minor adaptation of the defined error factors and refinement strategies based on the characteristics of the target dataset.

Overall, the results show that fine-grained diagnostic analysis combined with targeted adaptation strategies can substantially improve the reliability of HOI systems in educational environments with limited additional annotation. As future work, targeted ablation studies that systematically investigate the contribution of individual co-occurring error factors would provide deeper insight into error causation and further strengthen the proposed diagnostic framework. In addition, extending the framework with holistic error analysis [6, 29] would enable a quantitative assessment of how addressing specific error types influences overall performance, thereby supporting more principled, diagnosis-guided model refinement.

# 6. Acknowledgments

This research was partially supported by the Air Force Research Laboratory (AFRL) through the U.S. Army CCDC Soldier Center under Award No. W912CG2220001. The views expressed in this paper are those of the authors and do not reflect the official policy or position of the United States Government. No official endorsement should be inferred.

# References

[1] View-invariant object recognition using homography constraints | IEEE Conference Publication | IEEE Xplore. 3

[2] Hussain Ahmad Madni, Rao Muhammad Umer, and Gian Luca Foresti. Exploiting data diversity in multi-domain federated learning. *Machine Learning: Science and Technology*, 5(2):025041, 2024. 2

[3] T. S. Ashwin and Gautam Biswas. Identifying and Mitigating Algorithmic Bias in Student Emotional Analysis. In *Artificial Intelligence in Education*, pages 89–103, Cham, 2024. Springer Nature Switzerland. 3

[4] T. S. Ashwin, Nihar Sanda, Umesh Timalsina, and Gautam Biswas. Challenges of Applying Computer Vision for Emotion Detection in Educational Settings: A Study on Bias. In *Artificial Intelligence in Education*, pages 388–395, Cham, 2025. Springer Nature Switzerland. 2, 3

[5] Vijay Badrinarayanan, Alex Kendall, and Roberto Cipolla. SegNet: A Deep Convolutional Encoder-Decoder Architecture for Image Segmentation. *IEEE Transactions on Pattern Analysis and Machine Intelligence*, 39(12):2481–2495, 2017. 2

[6] Daniel Bolya, Sean Foley, James Hays, and Judy Hoffman. TIDE: A General Toolbox for Identifying Object Detection Errors, 2020. Version Number: 2. 2, 8

[7] Yu-Wei Chao, Zhan Wang, Yugeng He, Jiaxuan Wang, and Jia Deng. HICO: A Benchmark for Recognizing Human-Object Interactions in Images. In *2015 IEEE International Conference on Computer Vision (ICCV)*, pages 1017–1025, Santiago, Chile, 2015. IEEE. 2, 4, 5, 6

[8] Tianheng Cheng, Lin Song, Yixiao Ge, Wenyu Liu, Xinggang Wang, and Ying Shan. Yolo-world: Real-time open-vocabulary object detection. In *Proceedings of the IEEE/CVF conference on computer vision and pattern recognition*, pages 16901–16911, 2024. 6

[9] Joyce Fonteles, Clayton Cohn, Divya Mereddy, Ashwin T S, and Gautam Biswas. *Exploring Agentic Multimodal Late Fusion With LLMs for Embodied Learning*. 2025. 1

[10] Dan Hendrycks and Thomas Dietterich. Benchmarking Neural Network Robustness to Common Corruptions and Perturbations, 2019. arXiv:1903.12261 [cs]. 2, 3

[11] Derek Hoiem, Yodsawalai Chodpathumwan, and Qieyun Dai. Diagnosing Error in Object Detectors. In *Computer Vision – ECCV 2012*, pages 340–353. Springer Berlin Heidelberg, Berlin, Heidelberg, 2012. Series Title: Lecture Notes in Computer Science. 2

[12] Soonmin Hwang, Jaesik Park, Namil Kim, Yukyung Choi, and In So Kweon. Multispectral pedestrian detection: Benchmark dataset and baseline. In *2015 IEEE Conference on Computer Vision and Pattern Recognition (CVPR)*, pages 1037–1045, Boston, MA, USA, 2015. IEEE. 3

[13] Nichole Ingalls, David Zonies, Jeffrey A. Bailey, Kathleen D. Martin, Bart O. Iddins, Paul K. Carlton, Dennis Hanseman, Richard Branson, Warren Dorlac, and Jay Johannigman. A Review of the First 10 Years of Critical Care Aeromedical Transport During Operation Iraqi Freedom and Operation Enduring Freedom: The Importance of Evacuation Timing. *JAMA Surgery*, 149(8):807, 2014. 2

[14] Divya Mereddy Gautham Biswas Joyce Horn Fonteles, Clayton Cohn. Analyzing Embodied Learning in Classroom Settings: A Human-in-the-Loop AI Approach for Multimodal Learning Analytics. *Learning and Instruction*. 1

[15] Christoph Kamann and Carsten Rother. Benchmarking the Robustness of Semantic Segmentation Models with Respect to Common Corruptions. *International Journal of Computer Vision*, 129(2):462–483, 2021. 2

[16] Hiroyuki Kuromiya, Rwitajit Majumdar, and Hiroaki Ogata. Detecting Teachers' in-Classroom Interactions Using a Deep Learning Based Action Recognition Model. In *Artificial Intelligence in Education. Posters and Late Breaking Results, Workshops and Tutorials, Industry and Innovation Tracks, Practitioners' and Doctoral Consortium*, pages 379–382, Cham, 2022. Springer International Publishing. 3

[17] Shilong Liu, Zhaoyang Zeng, Tianhe Ren, Feng Li, Hao Zhang, Jie Yang, Qing Jiang, Chunyuan Li, Jianwei Yang, Hang Su, Jun Zhu, and Lei Zhang. Grounding dino: Marrying dino withnbsp;grounded pre-training fornbsp;open-set object detection. In *Computer Vision – ECCV 2024: 18th European Conference, Milan, Italy, September 29–October 4, 2024, Proceedings, Part XLVII*, page 38–55, Berlin, Heidelberg, 2024. Springer-Verlag. 6

[18] Yuen Peng Loh and Chee Seng Chan. Getting to know low-light images with the Exclusively Dark dataset. *Computer Vision and Image Understanding*, 178:30–42, 2019. 3

[19] Divya Mereddy, Marcos Quinones-Grueiro, Ashwin T. S, Eduardo Davalos, Gautam Biswas, Kent Etherton, Tyler Davis, Katelyn Kay, Jill Lear, and Benjamin Goldberg. Trainee Action Recognition through Interaction Analysis in CCATT Mixed-Reality Training, 2025. arXiv:2509.17888 [cs]. 1, 3, 6, 8

[20] S. Ning, L. Qiu, Y. Liu, and X. He. Hoiclip: Efficient knowledge transfer for hoi detection with vision-language models. In *Proceedings of the IEEE/CVF Conference on Computer Vision and Pattern Recognition (CVPR)*, pages 23507–23517, 2023. 1

[21] Connor Shorten and Taghi M. Khoshgoftaar. A survey on Image Data Augmentation for Deep Learning. *Journal of Big Data*, 6(1):60, 2019. 6

[22] C. Vatral, G. Biswas, C. Cohn, E. Davalos, and N. Mohammed. Using the dicot framework for integrated multimodal analysis in mixed-reality training environments. *Frontiers in Artificial Intelligence*, 5:941825, 2022. 1

[23] Tiancai Wang, Tong Yang, Martin Danelljan, Fahad Shahbaz Khan, Xiangyu Zhang, and Jian Sun. Learning Human-Object Interaction Detection Using Interaction Points. In *2020 IEEE/CVF Conference on Computer Vision and Pattern Recognition (CVPR)*, pages 4115–4124, Seattle, WA, USA, 2020. IEEE. 2

[24] Bingjie Xu, Yongkang Wong, Junnan Li, Qi Zhao, and Mohan S. Kankanhalli. Learning to Detect Human-Object Interactions With Knowledge. In *2019 IEEE/CVF Conference on Computer Vision and Pattern Recognition (CVPR)*, pages 2019–2028, Long Beach, CA, USA, 2019. IEEE. 6

[25] A. Zhang, Y. Liao, S. Liu, M. Lu, Y. Wang, C. Gao, and X. Li. Mining the benefits of two-stage and one-stage hoi detection. In *Advances in Neural Information Processing Systems (NeurIPS)*, pages 17209–17220, 2021. 1, 6

[26] Frederic Z. Zhang, Yuhui Yuan, Dylan Campbell, Zhuoyao Zhong, and Stephen Gould. Exploring Predicate Visual Context in Detecting of Human–Object Interactions. In *2023 IEEE/CVF International Conference on Computer Vision (ICCV)*, pages 10377–10387, Paris, France, 2023. IEEE. 1, 2, 4

[27] Kaiyang Zhou, Ziwei Liu, Yu Qiao, Tao Xiang, and Chen Change Loy. Domain Generalization: A Survey. *IEEE Transactions on Pattern Analysis and Machine Intelligence*, pages 1–20, 2022. 2

[28] Fangrui Zhu, Yiming Xie, Weidi Xie, and Huaizu Jiang. Diagnosing Human-object Interaction Detectors, 2023. arXiv:2308.08529 [cs]. 2

[29] Fangrui Zhu, Yiming Xie, Weidi Xie, and Huaizu Jiang. Diagnosing Human-Object Interaction Detectors. *International Journal of Computer Vision*, 133(4):2227–2244, 2025. 2, 3, 8

# Diagnosis of Human–Object Interaction Detectors for Real-World Educational Applications

Supplementary Material

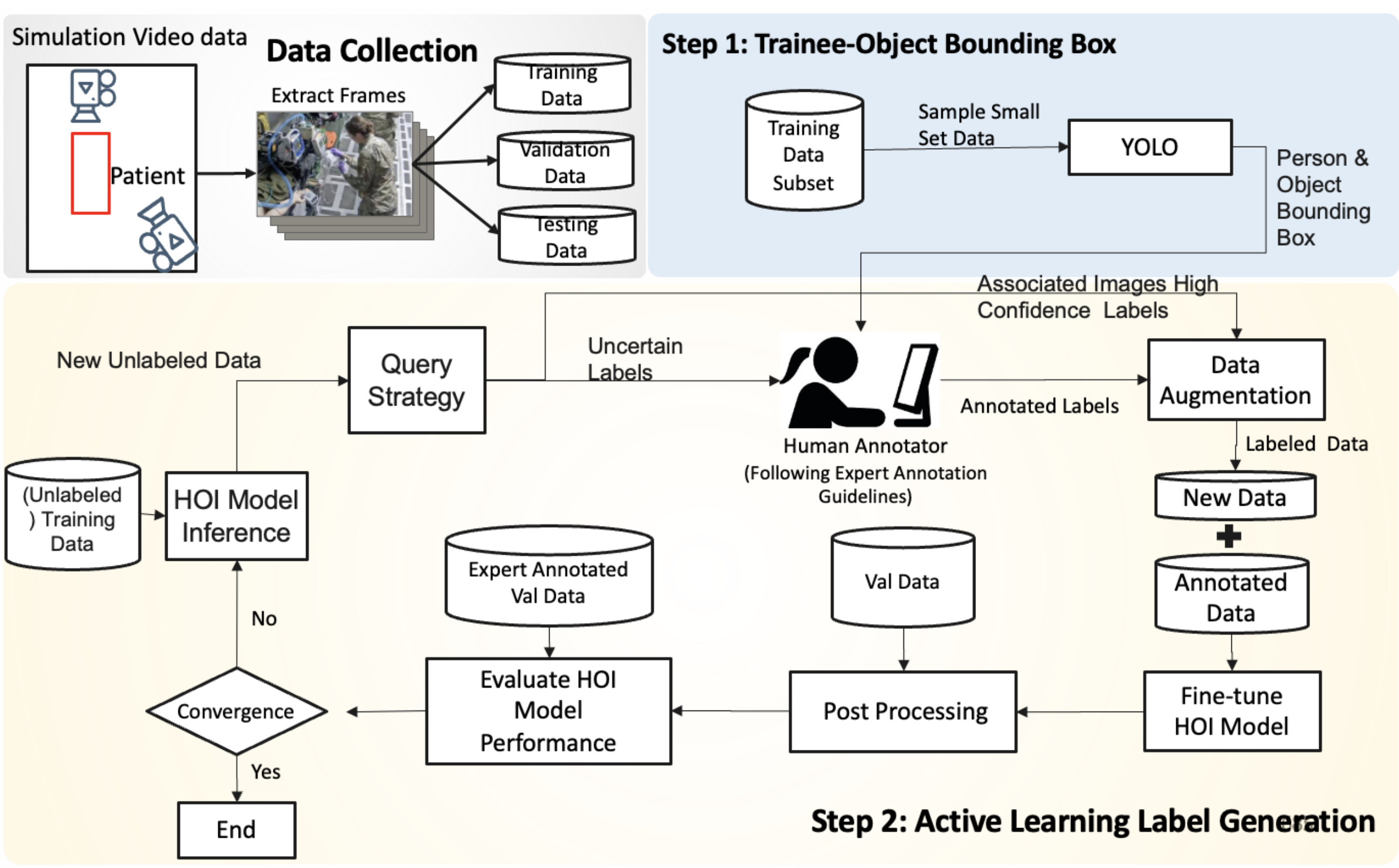

Figure 3. SAAL-HOI

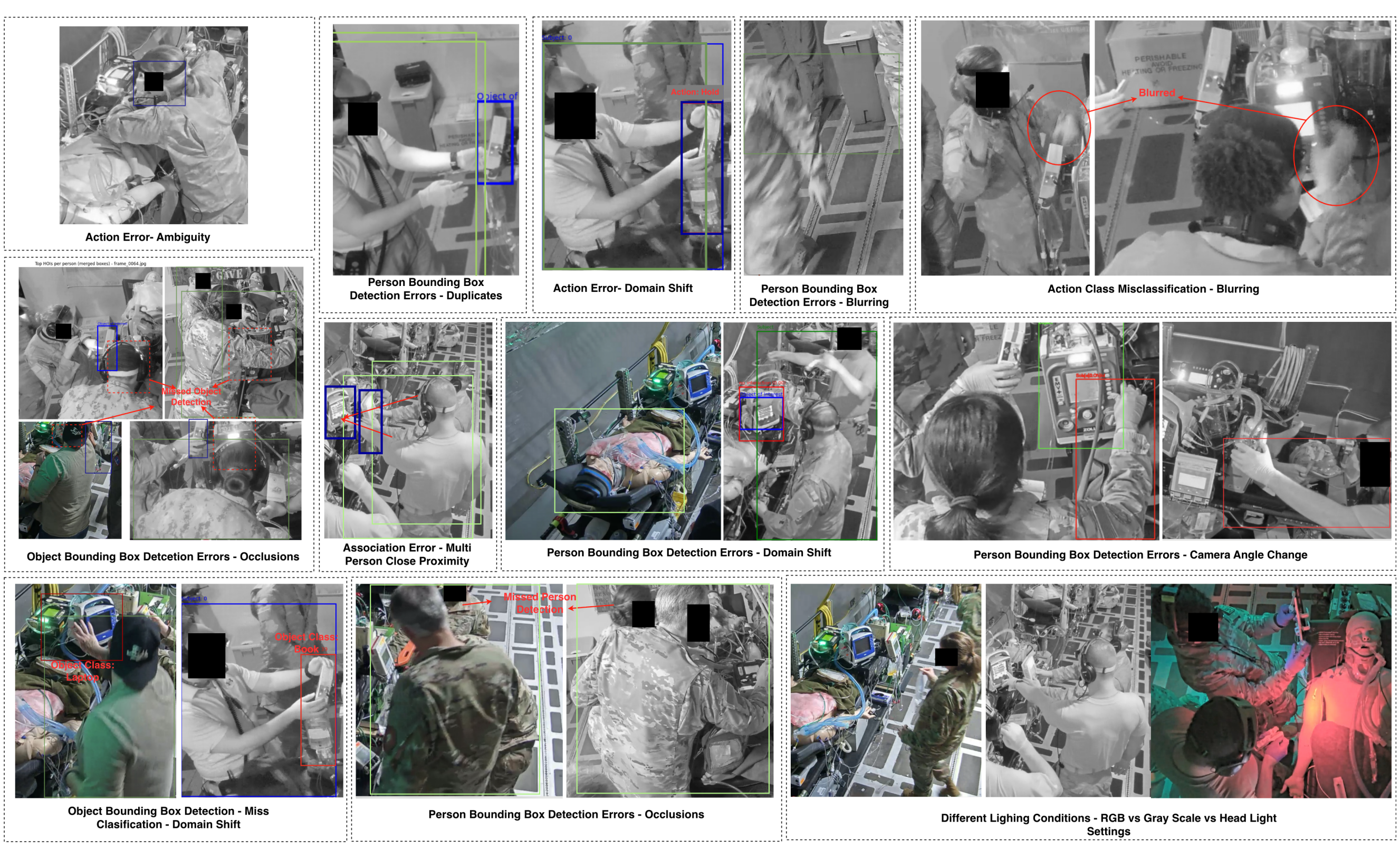

Figure 4. Human–Object Interaction (HOI) Error Categories: Qualitative Examples of CCATT-Specific Errors (Source: AFRL, Dayton)